\documentclass{article}
\usepackage[preprint,nonatbib]{neurips_arxiv}

\title{Infinite Class Mixup}

\usepackage{bm}
\usepackage{amsmath}
\usepackage{amssymb}
\usepackage{mathtools}
\usepackage{amsthm}
\usepackage{booktabs}
\usepackage{color,xcolor,colortbl}
\usepackage{ifthen}
\usepackage{wrapfig}
\definecolor{Gray}{gray}{0.9}
\newcommand{\logl}{\mathcal{L}}     %
\newcommand{\grad}[2][]{            %
    \ifthenelse{\equal{#1}{}}
    {\nabla_{\!#2}}
    {\nabla_{\!#2}{#1}}
}
\newcommand{\dirac}[1]{[\![{#1}]\!]}

\newcommand{\ty}{\tilde{\bm y}}     %
\newcommand{\tY}{\tilde{Y}}     %
\newcommand{\tx}{\tilde{\bm x}}     %
\newcommand{\tr}{\tilde{\bm r}}     %
\newcommand{\mc}[1][]{
    \ifthenelse{\equal{#1}{}}
    {C_{\ty}}
    {C_{\ty_{#1}}}
}

\newcommand*{\eg}{\textit{e.g.}\@\xspace}
\newcommand*{\ie}{\textit{i.e.}\@\xspace}

\newcommand{\supref}{Appendix~\ref{app:gradients}}
\newcommand{\camready}[1]{{\color{black}#1}}

\usepackage[colorlinks,urlcolor=blue,citecolor=red,bookmarks=false]{hyperref}
\usepackage{xspace}
\usepackage[sort,numbers]{bmvc2k_natbib}
\author{%
  Thomas Mensink\thanks{Paper accepted at BMVC 2023.\newline\mbox{}\hspace{5mm} Equal advising, writing, and coding. Order decided by coin flip.}\\
  Google Research\\
  \texttt{mensink@google.com}\\
  \And
  Pascal Mettes\footnotemark[1]\\
  University of Amsterdam\\
  \texttt{p.s.m.mettes@uva.nl}\\
}

\makeatletter
\DeclareRobustCommand\bmvaOneDot{\futurelet\@let@token\bmv@onedotaux}
\def\bmv@onedotaux{\ifx\@let@token.\else.\null\fi\xspace}
\makeatother

\def\eg{\emph{e.g}\bmvaOneDot}

\def\etal{\emph{et al}\bmvaOneDot}

\begin{document}

\maketitle
\begin{abstract}
Mixup is a widely adopted strategy for training deep networks, where additional samples are augmented by interpolating inputs and labels of training pairs. Mixup has shown to improve classification performance, network calibration, and out-of-distribution generalisation. While effective, a cornerstone of Mixup, namely that networks learn linear behaviour patterns between classes, is only indirectly enforced since the output interpolation is performed at the probability level. This paper seeks to address this limitation by mixing the classifiers directly instead of mixing the labels for each mixed pair. We propose to define the target of each augmented sample as a uniquely new classifier, whose parameters are a linear interpolation of the classifier vectors of the input pair. The space of all possible classifiers is continuous and spans all interpolations between classifier pairs. To make optimisation tractable, we propose a dual-contrastive Infinite Class Mixup loss, where we contrast the classifier of a mixed pair to both the classifiers and the predicted outputs of other mixed pairs in a batch. Infinite Class Mixup is generic in nature and applies to many variants of Mixup. Empirically, we show that it outperforms standard Mixup and variants such as RegMixup and Remix on balanced, long-tailed, and data-constrained benchmarks, highlighting its broad applicability. The code is available online at: \url{https://github.com/psmmettes/icm}.
\end{abstract}
\section{Introduction} 
There is a strong dependence between generalisation of deep networks and their access to rich and diverse samples for training~\cite{bommasani2021opportunities},
since deep neural networks tend to overfit to training sampled, or even memorise them \cite{goodfellow2016deep}.
Mixup forms a canonical approach to counteract this inclination~\cite{zhang2018mixup}. 
With Mixup, new samples are created by linearly interpolating input pairs and their corresponding ground truth outputs. By augmenting training pairs, a network is given insight into the linear transitions between classes, which helps to alleviate over-fitting.

Over the years, Mixup has shown to consistently improve down-stream performance for images~\cite{yun2019cutmix,baek2021gridmix}, videos~\cite{lu2022semantic}, point clouds~\cite{chen2020pointmixup}, graphs~\cite{wang2021mixup}, and more. Several works have observed that the improved performance of Mixup can be attributed to better network calibration~\cite{thulasidasan2019mixup} and out-of-manifold regularisation~\cite{guo2019mixup}. Due to its simplicity and strong empirical results, a wide range of Mixup variants have also been proposed \eg, to improve long-tailed recognition~\cite{chou2020remix}, semi-supervised learning~\cite{berthelot2019mixmatch}, and fairness~\cite{chuang2021fair}. At the core of training with Mixup is the following intuition:

\begin{center}
{\small
\textit{... Mixup extends the training distribution by incorporating the prior knowledge that linear interpolations of feature vectors should lead to linear interpolations of the associated targets.}
}\vspace{-3mm}
\flushright{\citet{zhang2018mixup}}
\end{center}

An assumption shared by all Mixup variants is that \emph{interpolation of the targets} should be at the probability level, which results in using cross-entropy losses with \emph{interpolated one-hot ground truth targets}.
For a Mixup interpolation between an image of a \emph{dog} and an image of a \emph{cat} with interpolation ratio $\lambda$,  the loss should enforce probability $\lambda$ for the \emph{dog} class and 1-$\lambda$ for the \emph{cat} class. Linear interpolation at the probability level, however, does not strictly imply linear classifier interpolation.
This paper argues instead that \emph{linear interpolations of input images should lead to linear interpolations of classifiers}.
From this view, for the interpolated image from the \emph{dog} and \emph{cat}, the target is a new (and unique) classifier, with its parameters given as the linear interpolation between the current \emph{dog} and \emph{cat} classifier weights. This allows us to directly enforce linear interpolation in the classifier space. 

Since the space of convex interpolations between all class pairs is continuous, there are an infinite number of Mixup classifiers to integrate over in a cross-entropy formulation, hence we name our proposed method \emph{Infinite Class Mixup}.
To make the optimisation tractable, we propose a dual-contrastive loss. For each mixed pair in a batch, we obtain a pair-specific classifier. We seek to optimise each pair towards their specific classifier and away from all other classifiers in the batch, resulting in an identity matrix with mixed pairs and their classifiers along the axes. We optimise with cross-entropy simultaneously over both axes, which have complementary gradient flows, and simply sum their losses.
\camready{These contrastive losses are not instantly applicable in standard Mixup, as it does not provide a direct setup to obtain positive and negative pairs. In Infinite Class Mixup, however, these pairs arise naturally because each example corresponds to a uniquely defined classifier, allowing contrastive losses between all unique examples and all classifiers in a batch.}

We show how Infinite Class Mixup can be integrated into different Mixup variants.
Empirically, we find that our Infinite Class formulation improves classification in standard, data-constrained, and imbalanced settings, outperforming both the conventional formulation and recent variants such as RegMixup~\cite{pinto2022regmixup} and Remix~\cite{chou2020remix}. 
Infinite Class Mixup does not require additional parameters and has similar computational cost compared to standard Mixup.
\section{Related work}

\paragraph{Mixup.}
Mixup as proposed by \citet{zhang2018mixup} is outlined for images by linearly interpolating image pairs for every pixel. Rather than interpolating on a global level, several works have proposed variants that interpolate images at a local level. For example, CutMix mixed images by masking part of one image onto a region of the other image, with the mask size equal to the interpolation ratio~\cite{yun2019cutmix}. Similarly, GridMix~\cite{baek2021gridmix} and RICAP~\cite{takahashi2018ricap} focus on a few subregions by dividing images into grids and randomly assigning image patches to the grids. PuzzleMix~\cite{kim2020puzzle}, co-Mixup~\cite{kim2021co}, and Attentive CutMix~\cite{walawalkar2020attentive} additionally include saliency or attention information to improve foreground selection in the image mixing, while StyleMix mixed both content and style for more visually coherent image mixes~\cite{hong2021stylemix}. AugMix performs Mixup between images and transformed versions~\cite{hendrycks2020augmix} and AlignMixup geometrically aligns images in feature space before mixing~\cite{venkataramanan2022alignmixup}. TokenMix mixes images at the token level for effective use in transformer models~\cite{liu2022tokenmix}. In Mixup without hesitation, the Mixup strategy is periodically turned off and on to speed up convergence and increase robustness to the interpolation hyperparameter~\cite{yu2021mixup}, while RegMixup combines the cross-entropy loss of the individual samples with the loss for mixed samples~\cite{pinto2022regmixup}. Manifold Mixup proposes to perform the interpolation on a latent manifold within the network, rather than the at the input-level~\cite{verma2019manifold}, which has shown to be effective for few-shot learning as well~\cite{mangla2020charting}. Other variants include TransMix \cite{chen2022transmix}, AutoMix \cite{liu2022automix}, and RecursiveMix \cite{yang2022recursivemix}.

Mixup and its variants focus on interpolating outputs at the probability level. This paper complements current Mixup literature by performing the output interpolation at the classifier-level instead of at the probability-level for improved down stream performance. \camready{For the inputs, we follow the original Mixup, so our proposed output interpolation could be combined with variants on input interpolation, such as Manifold Mixup~\cite{mangla2020charting,verma2019manifold}.}

\paragraph{Mixup in contrastive learning.}
Mixup variants have been proposed for contrastive learning. For example, 
Lee \etal~\cite{lee2021mix} and Ren \etal \cite{ren2022simple} perform Mixup in contrastive self-supervised learning with virtual labels, where one of the two views of an example is replaced with a Mixup variant of that view. In this paper, we use mixup to create an image and a classifier, and use these in a contrastive learning framework. 
Koshla \etal~\cite{khosla2020supervised} extend contrastive learning, to a supervised version, where multiple images from the same class can be used in a batch. We, on the other hand, use mixup to create unique pairs, so that each batch has only unique classes by construction. \camready{The concurrent Two-Way Loss \cite{kobayashi2023two} also performs contrastive learning on both axes of a sample-class matrix for multi-label classification. In contrast, we generate interpolated classes as per-sample targets and propose a dual-axis objective to improve general classification.}

\paragraph{Adapting Mixup to other tasks.}
Mixup outlines a general formulation to interpolate image samples. A wide range of works have therefore proposed task-specific extensions to Mixup. For long-tailed recognition problems, extensions include Remix~\cite{chou2020remix}, balanced Mixup~\cite{galdran2021balanced}, label occurrence-balanced Mixup~\cite{zhang2022label}, and dynamic Mixup~\cite{gao2022dynamic}, all of which bias the mixing coefficients toward minority classes. For out-of-distribution detection and robustness, extensions include adversarial vertex Mixup~\cite{lee2020adversarial} and Mixup during inference~\cite{pang2020mixup}. For semi-supervised data, Mixmatch provides labelled and unlabelled mixed images~\cite{berthelot2019mixmatch}, while Mix-and-Unmix in feature space improves semi-supervised object detection~\cite{kim2022mum}. 

Mixup has furthermore been extended to regression~\cite{yao2022c}, facial expression recognition~\cite{psaroudakis2022mixaugment}, fairness~\cite{chuang2021fair}, self-knowledge distillation~\cite{yang2022mixskd}, retrieval~\cite{patel2022recall}, domain adaptation~\cite{xu2022few,xu2020adversarial,zhou2022context}, COVID-19 detection in images~\cite{hou2021cmc}, zero-shot learning~\cite{xu2022generative}, and more. Our proposed Infinite Class Mixup is also viable to task-specific Mixup formulations, as highlighted by comparisons and experiments on long-tailed and data-constrained recognition tasks.

\paragraph{Mixup beyond images and videos.}
Mixup has also shown to be an effective learning strategy beyond interpolating pixels. PointMixup performs Mixup for point clouds by performing the interpolation between two training point clouds through optimal transport~\cite{chen2020pointmixup} and PointCutMix generalises CutMix to point clouds~\cite{zhang2022pointcutmix}. Other point cloud Mixup methods include Rigid SubSet Mixup \cite{lee2021regularization} and Point MixSwap \cite{umam2022point}.
Mixup has also been investigated for LiDAR \cite{xiao2022polarmix}, graphs \cite{wang2021mixup}, speaker verification~\cite{zhu2019mixup}, vision-language navigation~\cite{liu2021vision}, single-view 3D reconstruction~\cite{cheng2022pose}, and language processing \cite{kwon2022explainability,moysset2019manifold,sun2020mixup,zhao2021robust,wu2021mixup}.
We focus on Mixup for images, but our approach is generic and can be applied to many Mixup variants.

\section{Mixup with Infinite Classifiers}
Mixup uses interpolation between training examples to create an infinite amount of training data, which improves test generalisation, typically defined as follows:
\begin{equation}
\tx = \lambda \ \bm x_a + (1-\lambda) \ \bm x_b, \quad \quad\quad \quad \ty = \lambda \ \bm y_a + (1-\lambda) \ \bm y_b,
\end{equation}
where the mixup image $\tx$ is the result of interpolating input image $\bm x_a$ of class $A$ and $\bm x_b$ of class $B$, with interpolation ratio $\lambda$ (0$\leq$$\lambda$$\leq$1). The mixup target $\ty$ is the interpolation of the one-hot encoded ground truth vectors $\bm y_a$ and $\bm y_b$.
The interpolation ratio $\lambda$ is drawn from a parameterised $\textrm{Beta}(\alpha, \alpha)$ distribution. 

\begin{figure}[t]
    \centering
    \includegraphics[width=0.8\linewidth]{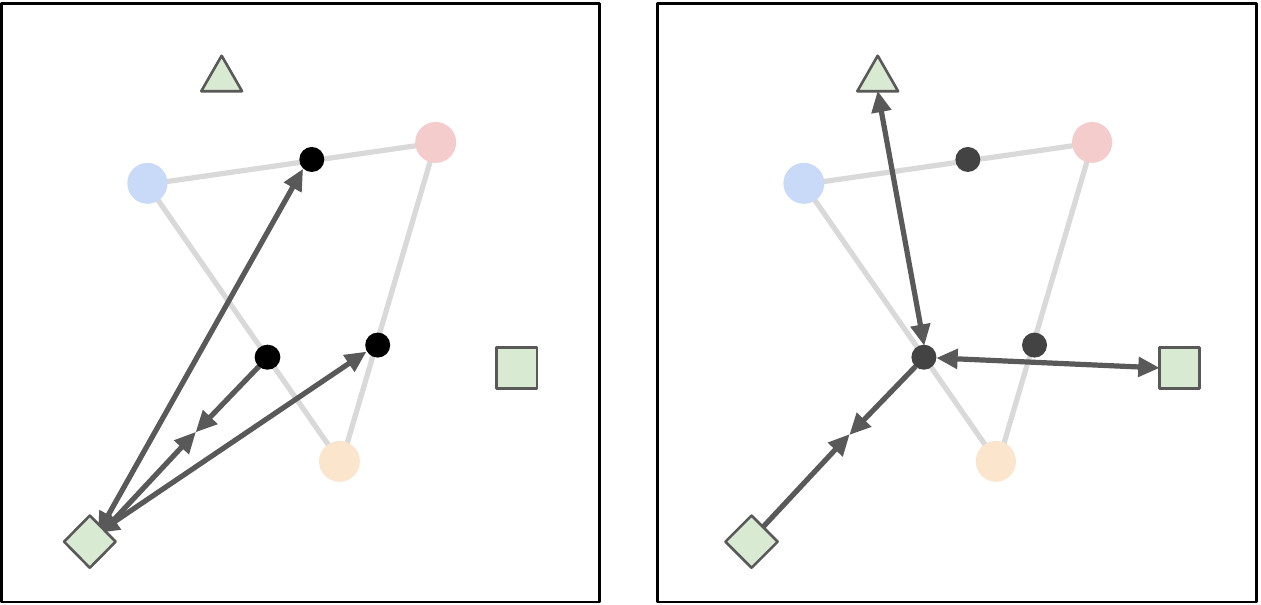}
    \caption{\textbf{Dual-contrastive learning in Infinite Class Mixup.} The blue, red, and yellow circles denote the original classifiers. The green shapes denote mixed samples, the black dots denote interpolated classifiers.
    Our contrastive loss is defined between a mixed example and all classes (\emph{left}) and between a class and all mixed examples in the same batch (\emph{right}).
    }
    \label{fig:contrastive}
\end{figure}

\paragraph{Interpolated classifiers.} 
The idea behind Infinite Class Mixup is that every mixup image $\tx$ corresponds to its own class $C_{\ty}$, 
defined as an interpolated class from the $C$ original classes using the mixup weights $\ty$. 
The class $C_{\ty}$ is fully specified by the $C$ original classes and the mixing weights $\ty$, since the mixing weights are continuous there are infinite many possible interpolated classes.

More formally, we construct the classifier weights $\bm w_{\ty}$, mixing the original classifier weights $\bm w_c$ proportionally to the Mixup weights:
\begin{equation}
    \bm w_{\ty} = \sum_{c} \ty_{c} W_{c} = W \ty,
\end{equation}
where $W \in \mathcal{R}^{D \times C}$ is a matrix with the classifier weights of the final layer of a deep neural network ($C$ number of original classes each having a $D$ dimensional weight vector), and $\ty \in \mathcal{R}^{C}$ is the vector with per-class mixing weights, \ie the contribution of each class to the Mixup example $\tx$.
Note that this formulation is rather generic, in the case that $\ty$ corresponds to a one-hot encoding of class $c$, the classifier weights $\bm w_{\ty} = \bm w_c$.

\paragraph{Interpolated class probability.}
We define the class probability of interpolated class $C_{\ty}$ similar to many softmax style classifiers:
\begin{align}
    p(C_{\ty} | \tx ) &\propto \exp( \tr^{\top} W \ty ), \quad\quad\quad\quad \textrm{using } \tr = f_{\theta}(\tx),\label{eq:class_prob}
\end{align}
where $\tr$ is the representation of image $\tx$ extracted from the penultimate layer of a deep convolutional network $f_{\theta}(\cdot)$.
To learn the values for the parameters $\{\theta,W\}$, we maximise the log-likelihood of the interpolated class prediction, as commonly used:
\begin{align}
    \logl
    &= \sum_{i} \log p( C_{\ty_i} | \tx_i). \label{eq:logloss}
\end{align}
We use the interpolated classes $C_{\ty}$ only during training, at test time we evaluate the original classes $C$ using the classifier weights $W$ and the network $f_{\theta}$.

In standard cross-entropy optimisation, the class probability of Equation \ref{eq:class_prob} is obtained through a normaliser $Z$ over all ground truth classes, given by $Z=\sum_{c'} \ \exp( \tr^{\top} \bm w_{c'})$. In our approach, the normaliser is given now over all possible interpolated classifiers, which forms a continuous space.
Thus, with infinite many classes this sum is intractable and hence we take a contrastive learning view where the normaliser $Z$ depends on the other examples in the batch. 
Below we introduce two contrastive variants.

\subsection{Contrastive learning of mixed samples}
Here, we offer a contrastive view on the cross entropy loss of Equation~\ref{eq:logloss}.
In many contrastive learning settings, positive pairs are formed by an image and its augmented version, and per batch negative pairs are sampled.
In Infinite Class Mixup, each Mixup image $\tx_i$ belongs to a unique class $\mc[i]$, this can also be seen as a positive pair which should be \textit{contracted}.
Interestingly enough, the sampling of negative pairs, which should be \textit{detracted}, can be done along two axis: across the different Mixup classes in the batch, or across the different images in the batch, see Figure~\ref{fig:contrastive}. Below we discuss both axes sequentially.

\paragraph{Contrasting classifiers.}
First we consider the setting where each image is compared to all classifiers, akin to a standard softmax classification network.
In this setting the positive pair $(\tx_i, \mc[i])$ is paired with negative pairs $(\tx_i, \mc[j])$, using the same image, but different interpolated classes from the same batch.
Then, the normaliser $Z$ in Eq.~\ref{eq:class_prob} is defined as follows:
\begin{align}
    Z_{\textbf{cc}} &= \sum_{j} \exp(\tr_i \ W^{\top} \ty_{j}), \label{eq:normalizer_cc}
\end{align}
which results in the following gradient for the classifier weights of class $c$:
\begin{align}
\grad[\logl_{\textbf{cc}}]{W_c}
        &= \sum_i \tr_i \left(\ty_{ic} - \sum_{j} p_{\textbf{cc}}(\mc[j] | \tx_i) \ty_{jc} \right),\label{eq:logl_cc}
\end{align}
where we use $\textbf{cc}$ to denote that we use contrastive classes, and $p_{\textbf{cc}}$ denotes the use of normaliser $Z_{\textbf{cc}}$ for the class probability of Eq.~\ref{eq:class_prob}.
We refer to \supref{} for the details on the derivation.

\paragraph{Contrasting interpolated images.}
Second we consider the setting where each classifier is compared to all interpolated images.
Instead of contrasting $(\tx_i, \mc[i])$ against pairs with the same image, we use pairs $(\tx_j, \mc[i])$ with the same classifier but different images.
This is equivalent in changing the normaliser to:
\begin{align}
    Z_{\textbf{ci}} &= \sum_j \exp(\tr_j \ W^{\top} \ty_i).
\end{align}
The gradient with respect to the weights $W_{c}$ of a class $c$ are then given by:
\begin{align}
    \grad[\logl_{\textbf{ci}}]{W_c}
        &= \sum_i \ty_{ic} \Bigl(\tr_i  - \sum_j p_{\textbf{ci}}(\mc[i] | \tx_j) \tr_j \Bigr),\label{eq:logl_ci}
\end{align}
where $\textbf{ci}$ denotes the use of contrastive interpolated images. This derivation is also given in \supref. \camready{Learning by contrasting other images is not directly applicable in standard Mixup variants, but is enabled with our Infinite Class perspective.}

\paragraph{Joint batch-level optimisation.} The two contrastive views visualised in Figure~\ref{fig:contrastive} allow for complementary optimisation. 
Yet the space of all possible pairs to contrast against remains infinite. 
As is common in contrastive learning~\cite{chen2020simple,khosla2020supervised}, we maximise the likelihood of Eq.~\ref{eq:class_prob} by contrasting against all other mixed samples in the same batch, resulting in:
\begin{align}
\logl
    &= \sum_{i}^{|B|} \log(p_{\textbf{cc}}( \mc[i] | \tx_i)) + \log(p_{\textbf{ci}}( \mc[i] | \tx_i)),\label{eq:batchloss}
\end{align}
where $|B|$ denotes the batch size.
This is implemented efficiently, by running a standard classification network up to the score (or logit) matrix $S \in \mathcal{R}^{B \times C}$ with the scores to the \emph{original dataset classes}.
Then we compute $\tilde{S} = S \tY^{\top}$, where $\tY$ is the $B \times C$ matrix of the stacked Mixup class contributions $\ty$, resulting in a $B \times B$ score matrix to the Mixup classes.
The contrastive loss (Eq.~\ref{eq:batchloss}) is then the cross entropy loss over both axes of the score matrix $\tilde{S}$, hence we denote class-axis as $p_{\textbf{cc}}$ and pair-axis as $p_{\textbf{ci}}$. In code, we call the cross entropy loss function with $\tilde{S}$ and with its transpose $\tilde{S}^{\top}$.

\begin{wrapfigure}{r}{0.4\textwidth}
\centering
%\vspace{-0.4cm}
\includegraphics[width=0.4\textwidth]{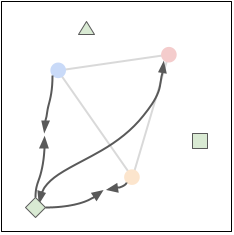}
\caption{\camready{In standard Mixup, samples are aligned with interpolated classes only at the normalized softmax level (curly lines) and cannot generalize to other contrastive axes.}}
\label{fig:vanilla}
\vspace{-1cm}
\end{wrapfigure}

\paragraph{Relation to Mixup.}
In Mixup, the gradient with respect to the classifier weights $W_{c}$ is:
\begin{align}
    \grad[\logl]{W_c}
        &= \sum_i \tr_i \Bigl(\ty_{ic} - p(c | \tx_i) \Bigr).
        \label{eq:logl_mixup}
\end{align}
The gradients of Mixup and Contrasting classifiers (Eq.~\ref{eq:logl_cc}) are similar, \ie both 
subtract the expected (predicted) class contribution $\mathbb{E}[\bm y_c]$ from the ground truth class contribution, resulting in: $\tr_i (\ty_{ic} - \mathbb{E}[\bm y_c])$, albeit both estimate the expectation differently.
\camready{Figure \ref{fig:vanilla} illustrates this difference: in standard Mixup, an example pulls towards the mixed classifiers, and retracts from the other classifiers, but the strength of each depends only on the post-softmax probabilities, \ie the curly lines, which are normalized scores instead of direct indiactors of classifier strength.}
The contrasting images loss (Eq.~\ref{eq:logl_ci}), however, 
rather looks at the expected class representation $\mathbb{E}[\tr]$ for class $\mc[i]$, resulting in $\ty_{ic} (\tr_i - \mathbb{E}[\tr])$.
Empirically we validate that the losses are complementary.
\newpage

\section{Experiments}

\paragraph{Implementation details.}
All experiments are done on ResNet \cite{he2016deep} and Wide ResNet \cite{zagoruyko2016wide} architectures. We train networks with Stochastic Gradient Descent  for 200 epochs with a learning rate of 0.1 and a decay by factor 0.2 after 50, 100, and 150 epochs, with momentum 0.9 and weight decay 5$e$-4. Unless specified otherwise, the batch size is set to 128. For Mixup and Remix, we set $\alpha$ to 0.2, for RegMixup we set $\alpha$ to 20. For Remix, we follow~\cite{chou2020remix} and set the interpolation threshold $\tau$ to 0.5 and the imbalance ratio threshold $\kappa$ to 3. All experiments are run three times and we show the mean accuracy results over the runs.

\subsection{Ablations and comparisons}

\begin{table}[t]
\centering
\resizebox{0.425\linewidth}{!}{%
\begin{tabular}[b]{@{}cc ccc@{}}
\toprule
 & & \multicolumn{3}{c}{\textbf{CIFAR-100}}\\
\multicolumn{2}{c}{contrastive axis} & \multicolumn{3}{c}{batch size}\\
 class-axis & pair-axis & 64 & 128 & 512\\
\midrule
$\checkmark$ & & 74.90 & 76.75 & 76.17\\
& $\checkmark$ & 75.38 & 77.62 & 76.09\\
\rowcolor{Gray}
$\checkmark$ & $\checkmark$ & \textbf{76.20} & \textbf{77.90} & \textbf{77.08}\\
\bottomrule
\end{tabular}
}
\hspace{0.75cm}
\resizebox{0.4\linewidth}{!}{%
\begin{tabular}[b]{@{}l cccc@{}}
\toprule
 & \multicolumn{4}{c}{\textbf{CIFAR-100}}\\
 & 100\% & 50\% & 25\% & 10\%\\
\midrule
No Mixup & 76.52 & 66.31 & 46.78 & 26.55\\
Mixup & 77.33 & 68.39 & 49.68 & 28.21\\
\midrule
\camready{IC-Mixup (c)} & 76.75 & 68.95 & 48.58 & \textbf{30.38}\\
\camready{IC-Mixup (p)} & 77.62 & \textbf{69.23} & \textbf{50.72} & 28.29\\
IC-Mixup & \textbf{77.90} & 68.89 & 50.48 & 30.19\\
\bottomrule
\end{tabular}
}
\caption{\textbf{Ablation studies on contrastive axes} (left) \textbf{and training size} (right) in Infinite Class Mixup. Consistently over the size of the batch and the fraction of the dataset size, combining both contrastive axes is effective and outperforms Mixup.}
\label{tab:batchsize}
\end{table}

\paragraph{Effect of dual-contrastive loss.} In Table~\ref{tab:batchsize} (left), we show the effect of the two contrastive axes in Infinite Class Mixup, as well as their summed loss. We compare the three variants on CIFAR-100 with a ResNet-34 architecture on three batch sizes.
We report the mean accuracy over three runs for all settings. For smaller batch sizes, we find that the pair-axis performs better, while we observe the reverse for large batch sizes. Across all three batch sizes, summing the losses of both contrastive axes is beneficial and improves the classification accuracy.
\camready{
We ran additional baselines where we contrast each interpolated example either to its two source classes, or to all original classes, instead of to all interpolated classes of the current batch.
These baselines do not outperform the proposed setup, and they do not directly allow for optimization over the pair-axis. Hence we follow the (more standard) contrastive approach.
}
We conclude that both contrastive axes of our Infinite Class Mixup are complementary and should be combined to improve down-stream performance.

\paragraph{Comparison to Mixup.} In Table~\ref{tab:batchsize} (right), we draw a comparison to the baseline Empirical Risk Minimisation without Mixup and to conventional Mixup. We report results for various training set sizes of CIFAR-100, where for each percentage we perform random stratified sampling per class to obtain a reduced training set.
When using the entire training set, Infinite Class Mixup obtains a mean accuracy of 77.90 compared to 77.33 for Mixup and 76.52 for the 
No Mixup baseline.
\camready{The improvements for Infinite Class Mixup also hold for all other training set sizes, however depending on the amount of data the class-axis only or the pair-axis only variant might perform slightly better. Overall, the fewer examples available, the more profitable it is to apply Infinite Class Mixup. We conclude from this experiment that Infinite Class Mixup is a viable alternative to Mixup for image classification.}

\begin{figure}[t]
\centering
\includegraphics[width=0.475\textwidth]{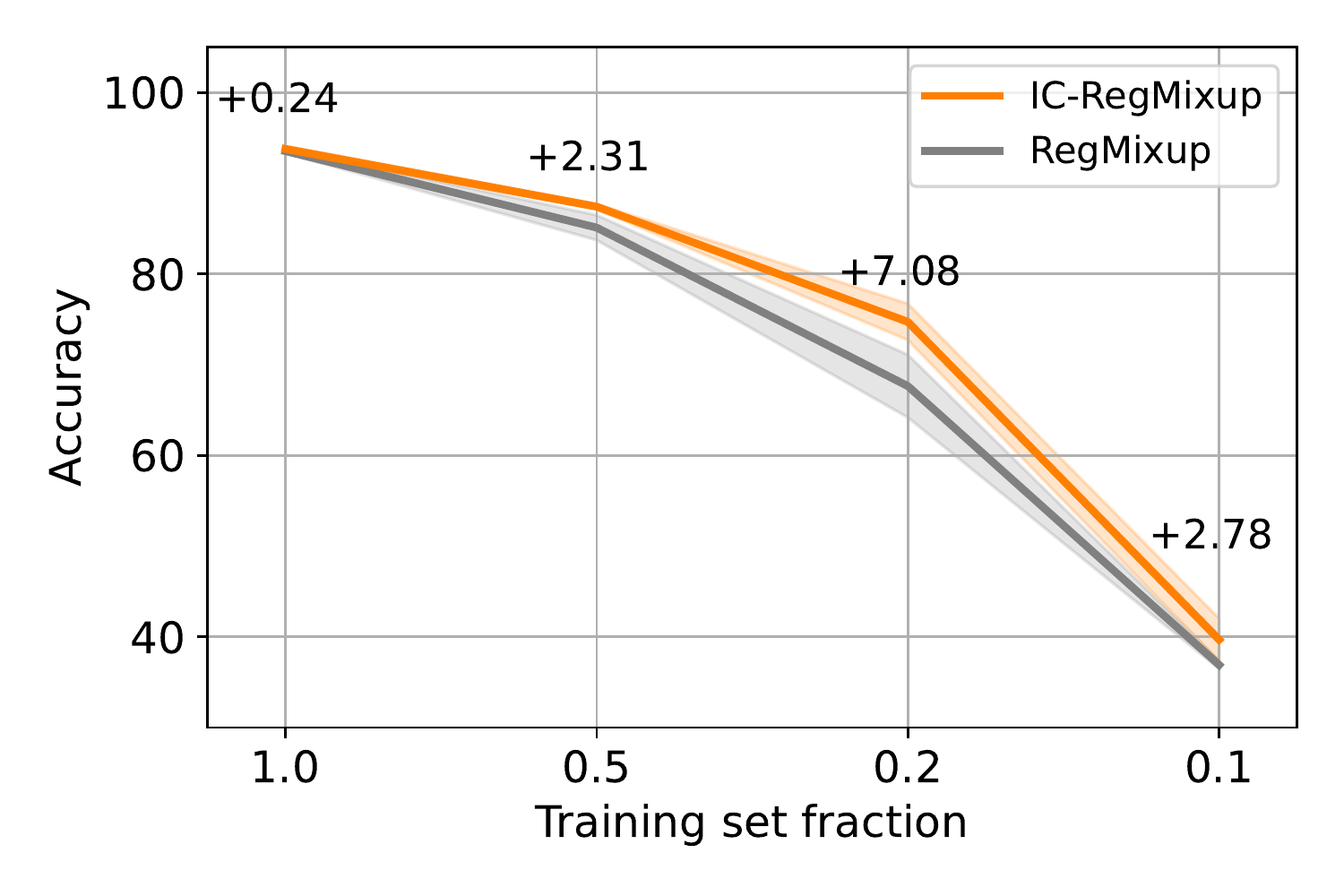}
\includegraphics[width=0.475\textwidth]{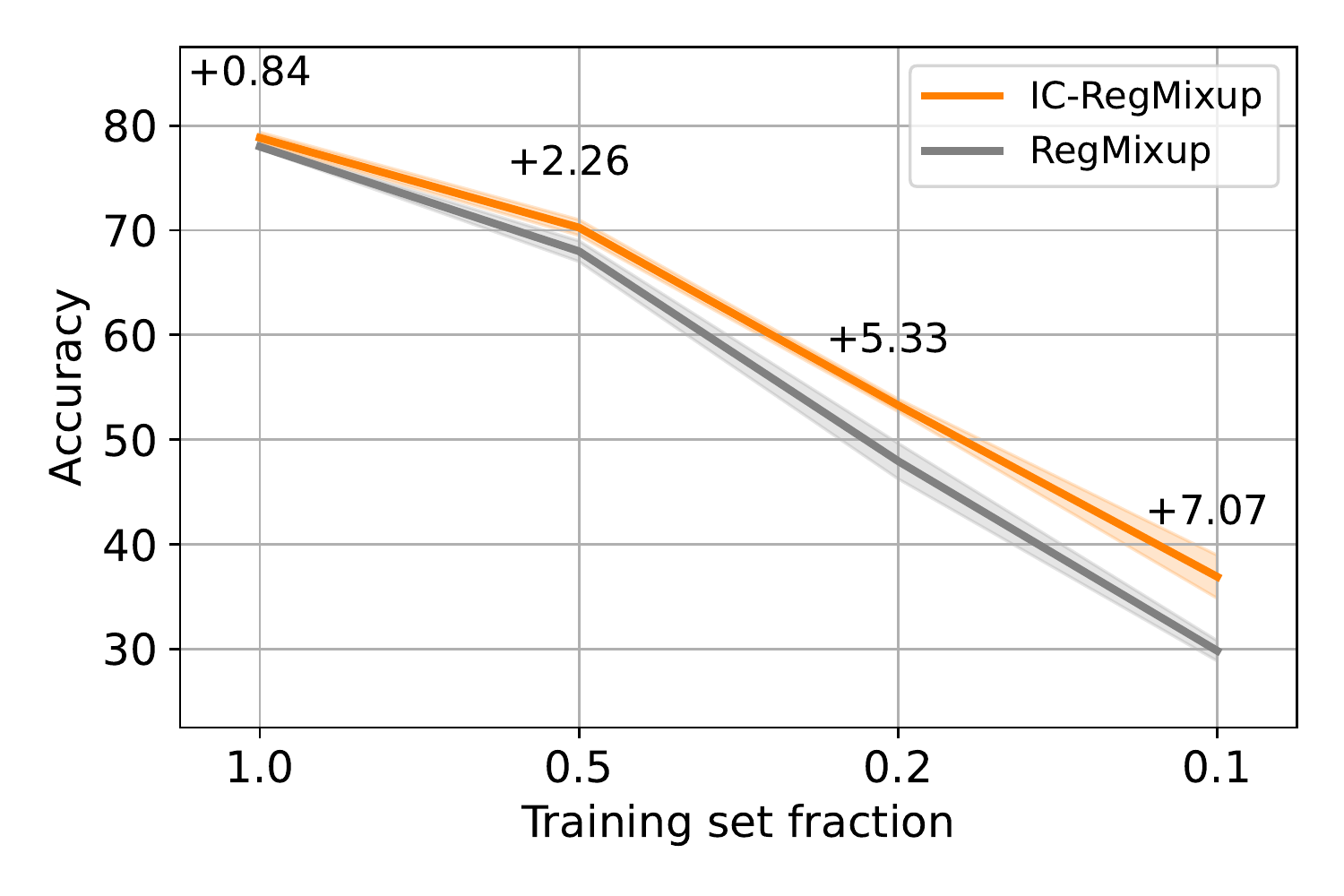}
\caption{\textbf{Comparing RegMixup~\cite{pinto2022regmixup} to Infinite Class RegMixup across various dataset sizes} on CIFAR-10 (left) and CIFAR-100 (right) with a ResNet-34 architecture. On both datasets, Infinite Class RegMixup outperforms RegMixup. For data-constrained settings with smaller training sets, mixing with infinite classes is preferred.}
\label{fig:regmixup-cifar}
\end{figure}

\paragraph{Comparison to RegMixup.} In Figure~\ref{fig:regmixup-cifar} we draw a comparison to the recently introduced RegMixup~\cite{pinto2022regmixup}. RegMixup provides a simple yet effective change to Mixup; rather than only training on mixed pairs, the mixed loss is added to a cross-entropy loss over the individual samples in a batch. This allows for exploring balanced interpolations between classes ($\alpha >> 1$), rather than sampling interpolations close to individual classes ($\alpha << 1$). %
We follow \citet{pinto2022regmixup} and use $\alpha = 20$.
On CIFAR-100 we find that Infinite Class RegMixup improves the mean accuracy on CIFAR-100 from 77.99 to 78.83 compared to RegMixup. As the training set size decreases, the difference in performance increases, up to an improvement of 7.07 p.p. when using 10\% of the training data, from 29.84 to 36.91. We observe similar improvements for CIFAR-10 and conclude that our Infinite Class formulation is also beneficial for RegMixup. Due to the improved results of RegMixup over Mixup overall, we recommend Infinite Class RegMixup for the best classification accuracy.

\subsection{Learning in constrained and long-tailed settings}

\paragraph{Data-constrained learning.} In Table~\ref{tab:other-compare} (left), we show experiments on ciFAIR-100 and ciFAIR-10, two datasets created to mimic learning with limited labels~\cite{brigato2021tune,barz2020we}. For this experiment, we train using the default hyperparameters akin to the other experiments. Overall, we find that mixing samples is an effective tool when dealing with limited samples. The best performance on both datasets is obtained using Infinite Class RegMixup.
In Table~\ref{tab:other-compare} (middle), we compare to the state-of-the-art on ciFAIR-10. \citet{brigato2021tune} have recently shown the big effect of precise hyperparameter tuning in such data-constrained settings, with top results on a tuned WideResNet-16-8 architecture. 
When using the same hyperparameters and and architecture, supplemented with Infinite Class RegMixup, the results improve from 58.22 to 61.84, reiterating the potential of our approach in data-constrained settings.

\paragraph{Long-tailed recognition.} For the imbalanced setting, we investigate on LT-CIFAR100 and LT-CIFAR10 for imbalanced ratios 0.1 and 0.01~\cite{cui2019class}. We do not adapt network training to the long-tailed domain and start from standard empirical risk minimization on the imbalanced training sets. We then incorporate Mixup~\cite{zhang2018mixup} and Remix, a variant of Mixup where the interpolation ratios of samples and classes is decoupled to account for long-tailed classes~\cite{chou2020remix}. Specifically, given interpolation ratio $\lambda$ and hyperparameters $\tau$ (interpolation threshold) and $\kappa$ (imbalance ratio threshold), the interpolation ratio $\lambda_y$ for the class probabilities is 0 if $n_i / n_j \geq \kappa$ and $\lambda < \kappa$, with $n_i$ the sample ratio of class $i$. The interpolation ratio is 1 if $n_i / n_j \leq 1 / \kappa$ and $1 - \lambda < \kappa$, and $\lambda$ otherwise.
Remix favours class assignments to long-tailed classes. In Table~\ref{tab:other-compare} (right), we report the results for empirical risk minimisation, Mixup, Remix, and our Infinite Class variants of Mixup and Remix. On both datasets, we find that our Infinite Class formulations improve over the standard formulations, especially for larger imbalance ratios. 
On LT-CIFAR100 (using 0.01 as imbalance ratio) the performance improves from 39.21 to 43.31 for Mixup and from 38.04 to 46.01 for Remix. We observe similar performance gains on LT-CIFAR10. 
Overall, while Remix performs on par or slightly below Mixup, Infinite Class Remix obtains the highest scores. 
We thus conclude that Infinite Class Remix is beneficial for imbalanced learning.

\begin{table}[t]
\centering
\resizebox{0.3\linewidth}{!}{%
\begin{tabular}[b]{@{}l cc@{}}
\toprule
& \textbf{ciFAIR-100} & \textbf{ciFAIR-10}\\% & \textbf{CUB-Birds}\\
\midrule
No Mixup & 41.96 & 45.58\\% & 00.00\\
\midrule
Mixup & 43.83 & 47.63\\% & 00.00\\
IC-Mixup & 43.11 & 49.03\\% & 00.00\\
\midrule
RegMixup & 47.55 & 53.17\\% & 00.00\\
\rowcolor{Gray}
IC-RegMixup & \textbf{47.67} & \textbf{55.54}\\% & 00.00\\
\bottomrule
\end{tabular}
}
\hspace{0.1cm}
\resizebox{0.3\linewidth}{!}{%
\begin{tabular}[b]{@{}l c@{}}
\toprule
& \textbf{ciFAIR-10}\\
\midrule
\citet{bietti2019kernel} & 51.03\\
\citet{oyallon2017scaling} & 54.21\\
\citet{kayhan2020translation} & 55.00\\
\citet{ulicny2019harmonic} & 56.50\\
\citet{kobayashi2021tvmf} & 57.50\\
\citet{brigato2021tune} & 58.22\\% \scriptsize{$\pm$ 0.90}\\
\rowcolor{Gray}
\textbf{+ IC-RegMixup} & \textbf{61.84}\\% \scriptsize{$\pm$ 0.42}\\
\bottomrule
\end{tabular}
}
\hspace{0.1cm}
\resizebox{0.3\linewidth}{!}{%
\begin{tabular}[b]{@{}l cc cc@{}}
\toprule
 & \multicolumn{2}{c}{\textbf{LT-CIFAR100}} & \multicolumn{2}{c}{\textbf{LT-CIFAR10}}\\
 & 0.1 & 0.01 & 0.1 & 0.01\\
\midrule
ERM & 58.54 & 37.44 & 88.63 & 71.87\\
\midrule
Mixup & 62.68 & 39.21 & 89.63 & 72.82\\
\rowcolor{Gray}
IC-Mixup & \textbf{64.30} & \textbf{43.31} & \textbf{89.89} & \textbf{76.81}\\
  & \scriptsize{+1.62} & \scriptsize{+4.10} & \scriptsize{+0.26} & \scriptsize{+3.99}\\
\midrule
Remix & 61.36 & 38.04 & 89.57 & 72.65\\
\rowcolor{Gray}
IC-Remix & \textbf{64.56} & \textbf{46.01} & \textbf{90.26} & \textbf{79.28}\\
  & \scriptsize{+3.20} & \scriptsize{+5.97} & \scriptsize{+0.67} & \scriptsize{+6.63}\\
\bottomrule
\end{tabular}
}
\caption{\textbf{Learning in constrained and long-tailed settings with Infinite Class Mixup.} Left: Infinite Class Mixup on top of RegMixup performs best on datasets with few samples per class. Middle: This formulation also outperforms other methods optimised for data-constrained settings. Right: Infinite Class Mixup also benefits long-tailed recognition.}
\label{tab:other-compare}
\end{table}

\subsection{What does Infinite Class Mixup learn differently?}
We have performed analyses on the impact of linear interpolation between classifiers for Mixup. We investigate (i) the example confidence as a function of the interpolation ratio and (ii) the difference in classifier dot products as a function of the interpolation ratio. For both analyses, we take ResNet-34 networks trained on CIFAR-100 and sample a single test image for each class. Then we sample all image pairs and construct interpolated images using $0\leq\lambda\leq1$ with step size 0.1, thus resulting in 99K interpolated images.

\begin{figure}[t]
\centering
\includegraphics[width=0.425\textwidth]{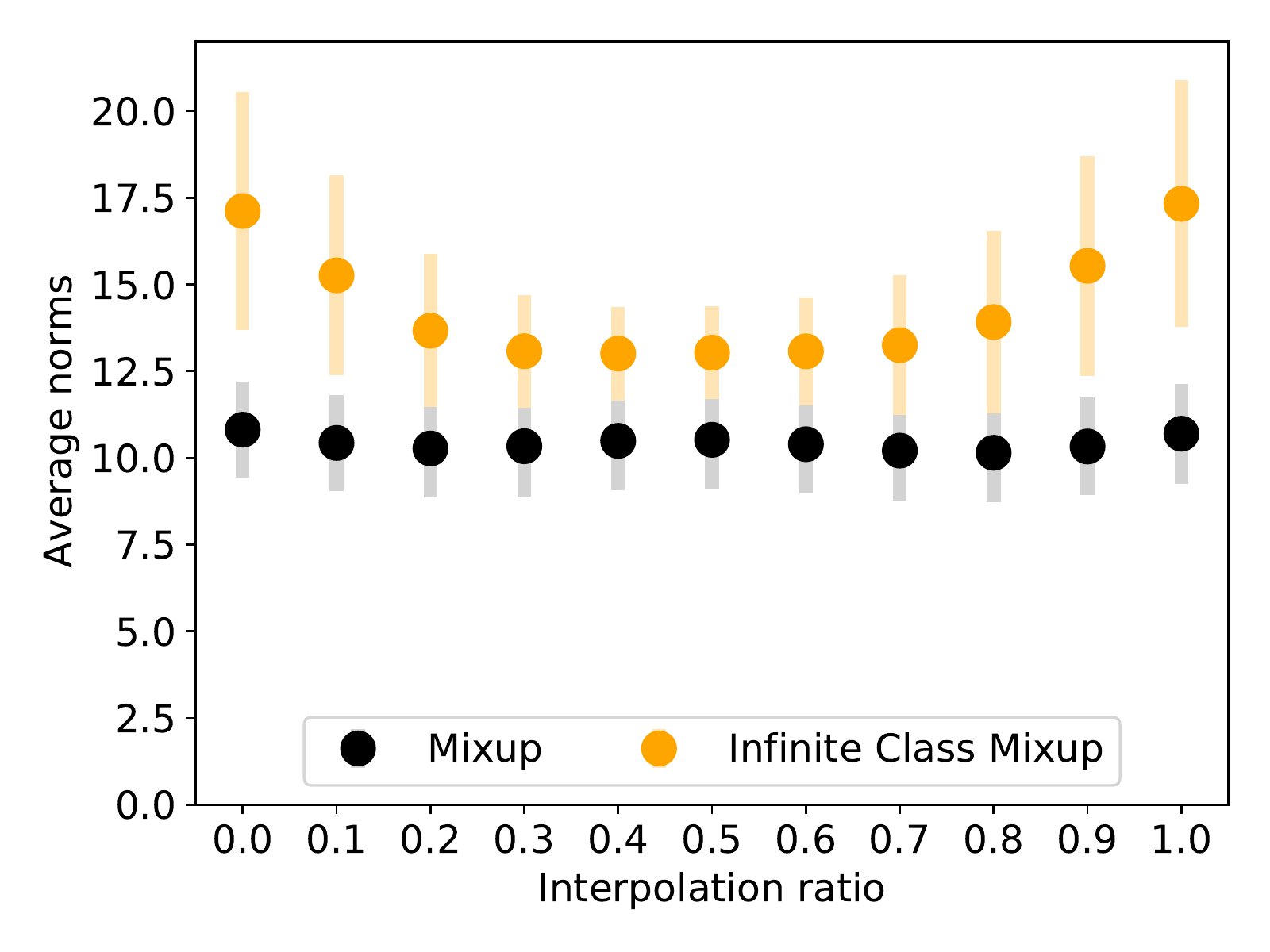}
\includegraphics[width=0.425\textwidth]{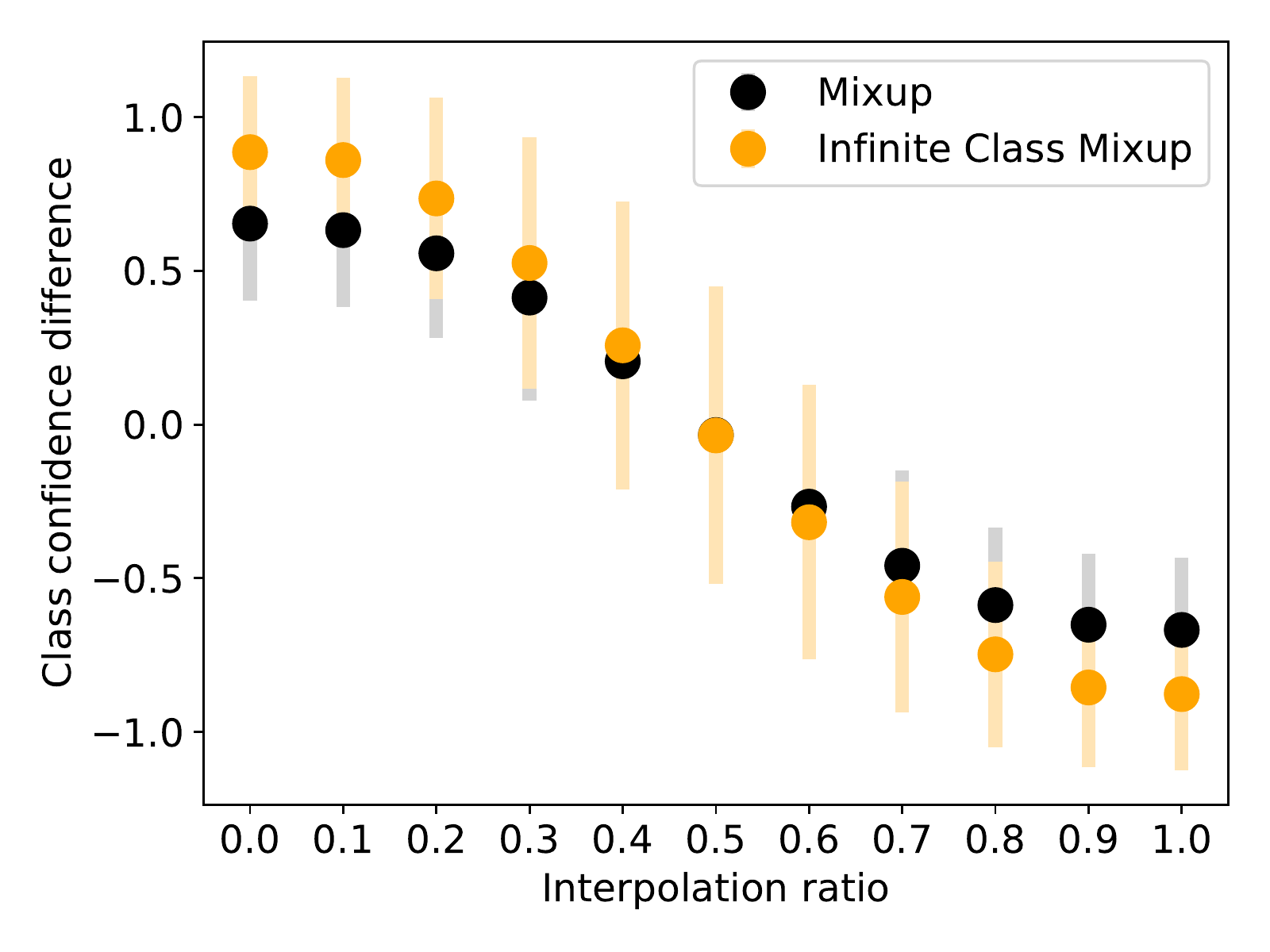}
\caption{\textbf{Understanding what Infinite Class Mixup learns.} Left: Infinite Class Mixup is on average more confident for un-interpolated images, which helps to differentiate between classes during testing. Right: The further the interpolation ratio is from 0.5, the bigger the difference in class confidence in Infinite Class Mixup. Our approach learns to better separate classes as a function of their inter-class ambiguity.}
\label{tab:analysis}
\end{figure}

\paragraph{Lower confidence for ambiguous interpolations.} In Figure~\ref{tab:analysis} (left), we show the mean confidence scores, computed by their class-independent squared norms, as a function of the interpolation ratio over all image pairs. We investigate networks trained with Mixup and with Infinite Class Mixup. As noted by \citet{liu2018decoupled}, the larger the norm, the more confident the prediction. When training with Mixup, interpolated samples have similar average norms compared to their original samples. For Infinite Class Mixup, however, the average norm is higher for uninterpolated images. In other words, Infinite Class Mixup is better equipped at differentiating ambiguous mixed samples and canonical unmixed samples.

\paragraph{Better differentiation between interpolated classes.} Finally, in Figure~\ref{tab:analysis} (right), we show the mean and stddev class confidence difference as a function of the interpolation ratio. For each image pair and interpolation ratio, we feed the mixed image to the network and compute the dot product with the classifiers of the labels of the pair. We take the dot product of the first class and subtract the dot product with the other class. %
Infinite Class Mixup shows a stronger relation between the interpolation ratio and the class confidence score, indicating that our formulation learns to better separate classes as a function of their inter-class ambiguity.
\section{Conclusions}
Mixup is a popular and effective algorithm for network training where images and their corresponding label vectors are linearly interpolated to generate new samples and diversify the training set. A linear interpolation of label vectors does however not ensure linear behaviour between classifiers as a function of the interpolation ratio. We introduce Infinite Class Mixup, where we interpolate images and their corresponding \emph{classifiers} directly. Each interpolated image is matched with a unique vector in classifier space, defined by a linear interpolation between the classifier vectors of the classes of the original image pair. We show how this setup can be optimised by contrasting simultaneously over both axes of the logit matrix between all image pairs and corresponding interpolated classes, improving classification in balanced, long-tailed, and data-constrained settings. 
\newpage
\appendix
\setcounter{equation}{0}
\renewcommand{\theequation}{\thesection.\arabic{equation}}
\section{Gradients of Mixup Variants.}
\label{app:gradients}
Here, we provide the gradient derivations of Mixup and Infinite Class Mixup.
All gradients are with respect to $W_c$, the linear classifier weights of one of the original classes $c$.

\subsection{Cross entropy gradient of standard Mixup}
For standard Mixup, the gradient of the cross entropy loss with respect to the classifier weights is given by:
\begin{align}
    \grad[\logl_{\textrm{mixup}}]{W_c} 
        &=  \grad{W_c} \sum_i \sum_{c'} \ty_{ic'} \log(p(c' | \tx_i))\label{eq:grad_logl_wrt_ck_detailed_start}\\
        &= \sum_i \sum_{c'} \ty_{ic'}\ \tr_i \ \left(\dirac{c'\!=\!c} - p(c | \tx_i) \right)\\
        &= \sum_i \tr_i \Bigl(\ty_{ic} - p(c | \tx_i) \Bigr)\label{eq:grad_logl_wrt_ck_detailed_end},
\end{align}
using: $\sum_{c'} \ty_{ic'} \dirac{c'\!=\!c} = \ty_{ic}$ and $\sum_{c'} \ty_{ic'} = 1$ in the last step, and in the first step the log derivative $\grad{} \log (f(x)) = \frac{1}{f(x)} \grad{} f(x)$ and the default softmax derivative:
\begin{align}
    \grad[p(c' | \tx_i)]{W_c} 
        &=  \grad{W_c} \frac{1}{Z} \exp(\tr^{\top}_i W_{c'}),\nonumber\\
        &= p(c' | \tx_i) \ \tr_i \left(\dirac{c'\!=\!c} - p(c | \tx_i) \right)\label{eq:grad_p_c_tx}.
\end{align}

\subsection{Cross entropy gradient of Infinite Class Mixup}
In the paper we propose two contrastive views to Infinite Class Mixup, one by contrasting classes, the other by contrasting pairs of interpolated images.
Their difference is reflected in the normalizer $Z$.
For both variants the following (partially defined) derivative of the likelihood holds:
\begin{align}
    \grad{W_c}p(\mc[i] | \tx_i)
        &= \frac{1}{Z} \grad{W_c} \exp(\tr_i^{\top} \ W \ty_i) - \frac{1}{Z^2} \exp(\tr_i^{\top} \ W \ty_i) \grad{W_c} Z\nonumber\\        
        &= p(\mc[i] | \tx_i) \left(\tr_i \ty_{ic} - \frac{1}{Z} \grad[Z]{W_c} \right),
\end{align}
which leads to:
\begin{align}
    \grad[\logl_{\textrm{inf}}]{W_c} 
        &= \sum_i \grad{W_c} \log p(\mc[i] | \tx_i)
        = \sum_i \frac{1}{p(\mc[i] | \tx_i)} \grad{W_c}p(\mc[i] | \tx_i),\nonumber\\
        &= \sum_i \left(\tr_i \ty_{ic} - \frac{1}{Z} \grad[Z]{W_c} \right).
\end{align}

\paragraph{Contrasting classes.}
The normalization factor of the contrasting classes variant is defined by varying the different classes in the batch:
\begin{align}
    Z_{\textbf{cc}} &= \sum_{j} \exp(\tr_i^{\top} \ W \ty_{j}),
\end{align}
which leads to the following gradient of $Z_{\textbf{cc}}$:
\begin{align}
    \frac{1}{Z_{\textbf{cc}}} \grad[Z_{\textbf{cc}}]{c}
        &= \frac{1}{Z_{\textbf{cc}}} \sum_{j} \grad{W_c} \exp(\tr_i^{\top} \ W \ty_{j})
        = \sum_{j} \frac{1}{Z_{\textbf{cc}}} \exp(\tr_i^{\top} \ W \ty_{j}) \tr_i \ty_{jc},\nonumber\\
        &= \sum_{j} p(\mc[j] | \tx_i) \tr_i \ty_{jc}.
\end{align}
Using this in the log-likelihood gradient $\grad[\logl_{\textbf{cc}}]{W_c}$, leads to:
\begin{align}
    \grad[\logl_{\textbf{cc}}]{W_c} 
        &= \sum_i \left(\tr_i \ty_{ic} - \sum_{j} p(\mc[j] | \tx_i) \tr_i \ty_{jc} \right)
        = \sum_i \tr_i \left(\ty_{ic} - \sum_j p(\mc[j] | \tx_i) \ty_{jc} \right).
\end{align}

\paragraph{Contrasting interpolated images.}
For the second variant, we obtain the normalization factor by contrasting to the embeddings of all other mixed pairs in the same batch as follows:
\begin{align}
    Z_{\textbf{ci}} &= \sum_{j} \exp(\tr_j^{\top} \ W \ty_{i}),
\end{align}
which leads to the following gradient of $Z_{\textbf{cc}}$:
\begin{align}
    \frac{1}{Z_{\textbf{ci}}} \grad[Z_{\textbf{ci}}]{c}
        &= \frac{1}{Z_{\textbf{ci}}} \sum_{j} \grad{W_c} \exp(\tr_j^{\top} W \ty_{i})
        = \sum_j \frac{1}{Z_{\textbf{ci}}} \exp(\tr_j^{\top} W \ty_i) \tr_j \ty_{ic},\nonumber\\
        &= \sum_j p(\mc[i] | \tx_j) \tr_j \ty_{ic}.\label{eq:grad_1_Z_ck}        
\end{align}
When plugged into the gradient $\grad[\logl_{\textbf{ci}}]{W_c}$, we obtain the following gradient derivation:
\begin{align}
    \grad[\logl_{\textbf{ci}}]{W_c} 
        &= \sum_i \left(\tr_i \ty_{ic} - \sum_j p(\mc[i] | \tx_j) \tr_j \ty_{ic} \right)
        = \sum_i \ty_{ic} \left(\tr_i - \sum_j p(\mc[i] | \tx_j) \tr_j \right).
\end{align}

\bibliography{egbib}
\end{document}